\documentclass[lettersize,journal]{IEEEtran}
\usepackage{amsmath,amsfonts}
\usepackage{algorithmic}
\usepackage{algorithm}
\usepackage{array}
\usepackage[caption=false,font=normalsize,labelfont=sf,textfont=sf]{subfig}
\usepackage{textcomp}
\usepackage{stfloats}
\usepackage{url}
\usepackage{verbatim}
\usepackage{graphicx}
\usepackage{cite}
\usepackage{multirow}
\usepackage[table]{xcolor}
\usepackage{color}

\newcommand{\blue}[1]{\textcolor{blue}{#1}}
\usepackage{float}
\usepackage{booktabs}
\usepackage[colorlinks,
            linkcolor=blue,   
            anchorcolor=blue,
            citecolor=blue,    
            ]{hyperref}

\usepackage{pifont}
\begin{document}

\title{Weakly-supervised Contrastive Learning with Quantity Prompts for Moving Infrared Small Target Detection}


\author{Weiwei Duan, Luping Ji$^\ast$, \IEEEmembership{Member, IEEE}, Shengjia Chen, Sicheng Zhu, Jianghong Huang, Mao Ye, \IEEEmembership{Senior Member, IEEE}
\thanks{$^\ast$Corresponding author: Luping Ji.}
\thanks{This work is supported by the National Natural Science Foundation of China (NSFC) under Grant No. 62476049, and the Aeronautical Science Foundation of China (ASFC) under Grant No. 2022Z071080006.}
\thanks{Weiwei Duan, Luping Ji, Shengjia Chen, Sicheng Zhu, Jianghong Huang and Mao Ye are with the School of Computer Science and Engineering, University of Electronic Science and Technology of China, Chengdu 611731, China (e-mail: dww@std.uestc.edu.cn;
	jiluping@uestc.edu.cn; 
    shengjiachen@std.uestc.edu.cn;
    cvlab.uestc@gmail.com)} }

\markboth{Journal of \LaTeX\ Class Files,~Vol.~$\ast\ast$, No.~$\ast\ast$, June~2025}%
{Shell \MakeLowercase{\textit{et al.}}: A Sample Article Using IEEEtran.cls for IEEE Journals}


\maketitle

\begin{abstract}
Different from general object detection, moving infrared small target detection faces huge challenges due to tiny target size and weak background contrast.
Currently, most existing methods are fully-supervised, heavily
relying on a large number of manual target-wise annotations.
However, manually annotating video sequences is often expensive and time-consuming, especially for low-quality infrared frame images. Inspired by general object detection, non-fully supervised strategies ($e.g.$, weakly supervised) are believed to be potential in reducing annotation requirements.
To break through traditional fully-supervised frameworks, as the first exploration work, this paper proposes a new weakly-supervised contrastive learning (WeCoL) scheme, only requires simple target quantity prompts during model training.
Specifically, 
in our scheme, based on the pretrained segment anything model (SAM), a potential target mining strategy is designed to integrate target activation maps and multi-frame energy accumulation.  
Besides, contrastive learning is adopted to further improve the reliability of pseudo-labels, by calculating the similarity between positive and negative samples in feature subspace.
Moreover, we propose a long-short term motion-aware learning scheme to simultaneously model the local motion patterns and global motion trajectory of small targets.
The extensive experiments on two public datasets (DAUB and ITSDT-15K) verify that our weakly-supervised scheme could often outperform early fully-supervised methods. Even, its performance could reach over 90\% of state-of-the-art (SOTA) fully-supervised ones.
Source codes are available at \blue{\url{https://github.com/UESTC-nnLab/WeCoL}}. 
\end{abstract}

\begin{IEEEkeywords}
Weakly-supervised Learning, Target Quantity Prompts, Contrastive Learning, Moving Infrared Small Target Detection, Segment Anything Model
\end{IEEEkeywords}

\section{Introduction}
\IEEEPARstart{I}{nfrared} small target detection (ISTD) has the advantages of being independent of external lighting and working in almost all weather conditions \cite{pengmixed}. 
It is indispensable in various areas, such as 
early invasion warning, remote sensing and rescue missions \cite{duan2025semi, tmp}.
As a significant research branch of object detection, it has garnered more and more attention over the past decades \cite{PSTNN, AGPCNet, DTUM}. 

Compared to general object detection \cite{survey_tnn}, ISTD often presents two significant challenges. One is that due to long imaging distance, infrared targets are typically small and dim, often lacking distinct visual features, which is more likely to lead to missed detections \cite{stdmanet}.
The other is that the cluttered backgrounds with lots of noise could often result in false detections \cite{small_tnn}.
Therefore, accurately locating, detecting, and recognizing moving small targets in infrared background images and videos is often challenging and meaningful.

In early stages, many model-driven methods were firstly proposed, treating ISTD as a filtering and target enhancement problem, $e.g.$, MaxMean \cite{MaxMean}, Top-Hat \cite{Tophat} and 5D-STFC \cite{5D-STFC}.
This type of methods often depends heavily on the prior knowledge of infrared images, and hand-crafted features, lacking learning ability. 
They usually have high false alarm rates in real-world applications.
\begin{figure}[t]
    \centering
\includegraphics[width=\linewidth]{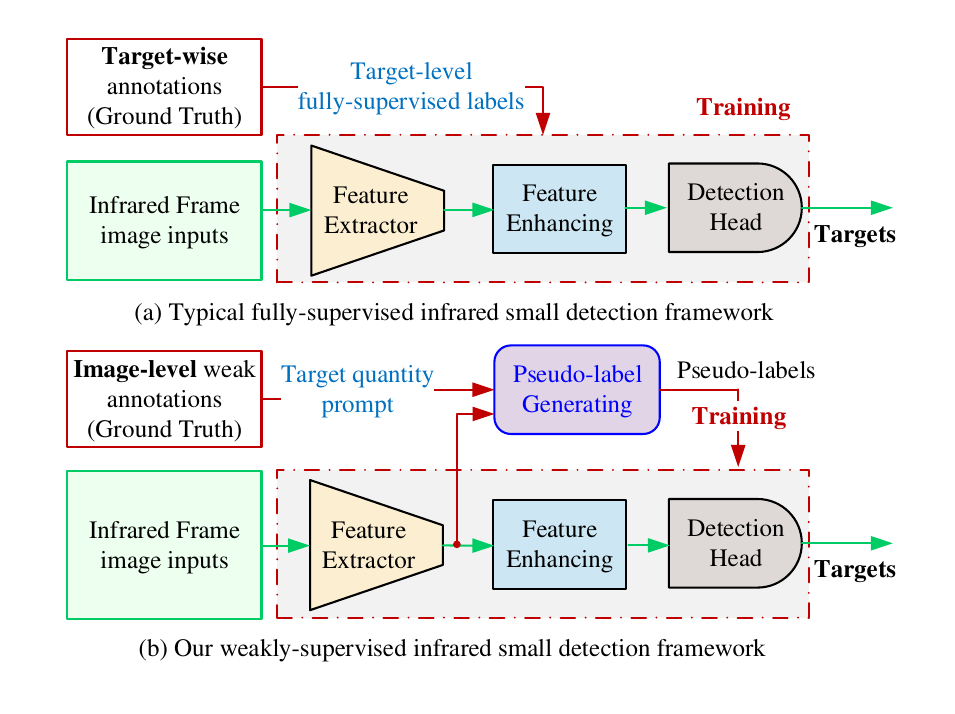}
    \caption{The comparisons between typical fully-supervised scheme and our weakly-supervised scheme. The green arrows in our scheme denote real inference pipleline.}
    \label{fig:intro}
\end{figure}

To address the apparent drawbacks of model-driven methods, in recent years, many data-driven schemes have been proposed \cite{dnanet}, with the development of deep learning. 
Depending on the number of input frames, data-driven methods can be further categorized into \emph{single-frame} and \emph{multi-frame}-based approaches.
Typical single-frame schemes only rely on the visual features of a single image, with low complexity and high inference speed, $e.g.$, 
MSHNet \cite{MSHNet} and LSKNet \cite{wu2024lsk}.
However, due to the ignorance of temporal information between consecutive frames, they are often less effective in complex scenarios.
In contrast, existing multi-frame ISTD (MISTD) methods could extract target features from both visual and temporal patterns, $e.g.$, STDMANet \cite{stdmanet}, ST-Trans \cite{sttrans} and Tridos \cite{tridos}. They often outperform the pure visual features of single-frame detection schemes.

Totally observed, almost all existing data-driven methods rely on fully-supervised learning, requiring sufficient target-wise annotations, as shown in Figure \ref{fig:intro} (a). Once no sufficient annotations are available, they will suffer severe performance degradation. 
Besides, target-wise annotation is extremely high-cost, especially for moving infrared small targets. 
Inspired by general object detection \cite{zhang2021weakly}, to address this annotation problem, weakly-supervised scheme is usually believed to be one of the most promising solutions.

At present, weakly-supervised learning has been widely investigated in traditional object detection, $e.g.$, WSDDN \cite{WSDDN}, OCIR \cite{OCIR} and ESFL \cite{ESFL}.  
They usually utilize class activation maps (CAM) \cite{PLE} or traditional region proposal algorithms \cite{selective} to generate pseudo-labels based on the strong visual features of general objects. 
However, for infrared small targets, strong visual features could not be effectively captured.
Moreover, almost all existing weakly-supervised object detection methods only process an independent image in a single-frame way, ignoring the rich motion features in consecutive frames.
Besides, compared to general objects, noisy pseudo-labels have more serious effects on small targets, and wrong samples are more difficult to correct in training.
Therefore, our \textbf{primitive motivation} is to break through traditional high-cost target-wise annotations and present a new weakly-supervised learning scheme, especially for moving infrared small targets.

A key challenge of weakly-supervised MISTD is how to obtain reliable pseudo-labels to supervise detector training. 
This is because, in traditional object detection, even if some pseudo-labels are wrong, general objects often have extra visual information to compensate.
In contrast, due to the tiny infrared target size, once a large number of wrong samples are mixed, the model will overfit very easily.
Moreover, previous weakly-supervised methods \cite{WSDDN,CBL} often employ a fixed threshold or non-maximum suppression to filter pseudo-labels. This could often increase the probability of selecting wrong pseudo-labels. 
As such, a natural question arises: \textbf{\emph{Can we know in advance how many targets are in a frame to guide detector training ?}} 
The target quantity prompts could be used as soft constraints to select high-quality pseudo-labels. 

Inspired by the above analysis, we propose a \textbf{W}eakly-supervised \textbf{Co}ntrastive \textbf{L}earning framework with quantity prompts (\textbf{WeCoL}) for MISTD, as shown in Figure \ref{fig:intro} (b). 
Different from traditional fully-supervised schemes, it doesn't need high-cost target-wise annotations, by utilizing SAM with adaptive point prompts to mine potential targets. 
Besides, it adopts contrastive learning between positive and negative samples to select high-quality pseudo-labels with quantity prompts guiding. 
The experiments on two public benchmarks show that our weakly-supervised scheme could achieve over 90\% performance to fully-supervised SOTA methods.
In summary, our primary contributions are as follows:

(I) Breaking through the traditional fully-supervised training schemes requiring target-wise annotations, the first weakly-supervised contrastive learning framework only with target quantity prompts is proposed in this paper.

(II) A potential target mining strategy to generate initial pseudo-labels is designed, integrating the activation maps and energy accumulation by classic SAM.

(III) A pseudo-label contrastive learning strategy with infrared target quantity prompt guiding is proposed to obtain high-quality positive samples for detector training. 

(IV) To enhance target features, a long-short term motion-aware learning approach is developed to jointly model the local and global motion features of infrared small targets.

\section{Related Work}
\subsection{Moving Infrared Small Target Detection}
According to the number of input images, ISTD could often be divided into single-frame and multi-frame schemes \cite{sstnet}. They could be further classified as model-driven and data-driven methods.

Representative single-frame model-driven methods include human visual system (HVS)-based detection techniques, $e.g.$, local contrast measure (LCM) \cite{lcm} and WSLCM \cite{wslcm}. Additionally, some data structure-based approaches separate targets using the special characteristics of infrared images \cite{PSTNN}.
With the development of deep learning, data-driven ones have powerful adaptive learning abilities on numerous labeled samples, establishing them as a dominant paradigm \cite{ACM, AGPCNet}. 
For example, DNANet \cite{dnanet} proposes a dense nested interactive module to avoid target loss in deep layers. 
Moreover, MSHNet \cite{MSHNet} designs a scale and location-sensitive loss to help detectors localize and distinguish targets precisely. 
RPCANet \cite{rpcanet} introduces a deep unfolding network by robust principle component analysis. 
However, these single-frame methods primarily focus on still images, ignoring the significant motion patterns in moving infrared small targets.

Unlike single-frame images, MISTD could provide rich motion information, offering great potential to improve detection performance \cite{sstnet}. 
Traditional MISTD algorithms often use methods like energy accumulation \cite{zhou2021background} and motion estimation \cite{optical} to detect targets by calculating the difference between consecutive frames.
Recently, some tensor-optimized methods \cite{4Dtensor, 5D-STFC} have shown remarkable results. They often separate targets by employing the special characteristics of infrared images.
Despite advancements in these traditional methods, they often struggle to effectively handle complex real-world scenarios \cite{tmp}. 

With the development of deep neural networks, data-driven multi-frame methods have emerged as a promising alternative.
For example, ST-Trans \cite{sttrans} designs a spatio-temporal transformer to extract motion dependencies between successive frames. 
Tridos \cite{tridos} expands the pattern of feature learning from a spatio-temporal-frequency domain.
Recently, DTUM \cite{DTUM} proposes a direction-coded temporal U-shape module to distinguish the motion of targets from different directions. 
However, these methods depend heavily on a large number of expensive manual target-wise annotations.  

\subsection{Weakly-supervised Object Detection}
Weakly-supervised Object Detection (WSOD) expects to train detectors with image-level annotations to reduce the burden of manual labeling \cite{CBL,zhu2024weaksam}.
In early stages, most works utilize the Multiple Instance Learning (MIL) strategy to transform WSOD into a multi-class classification task, $e.g.$, WSDDN \cite{WSDDN} and OCIR \cite{OCIR}.
The later works aim to improve WSOD performance from different perspectives.
For example, 
CBL \cite{CBL} devises the cyclic-bootstrap labeling strategy to optimize multiple instance detection networks with rank information by a weighted ensemble teacher.
Furthermore, ESFL \cite{ESFL} proposes enhanced spatial feature learning to employ the multiple kernels in a single pooling layer to handle multiscale objects.
Recently,  PFPM \cite{PFPM} designs a progressive frame-proposal mining framework to exploit discriminative proposals in a coarse-to-fine way for weakly-supervised video object detection.
WeakSAM \cite{zhu2024weaksam} introduces a weakly-supervised instance-level recognition pipeline to address incorrect pseudo-labeling.   
Moreover, PLE \cite{PLE} uses self-supervised vision transformers to enhance the localization ability and completeness of pseudo-labels.
However, existing WSOD methods often concern general objects in still images, rendering them ineffective in challenging MISTD scenarios.
\begin{figure*}[t]
    \centering
\includegraphics[width=0.92\linewidth]{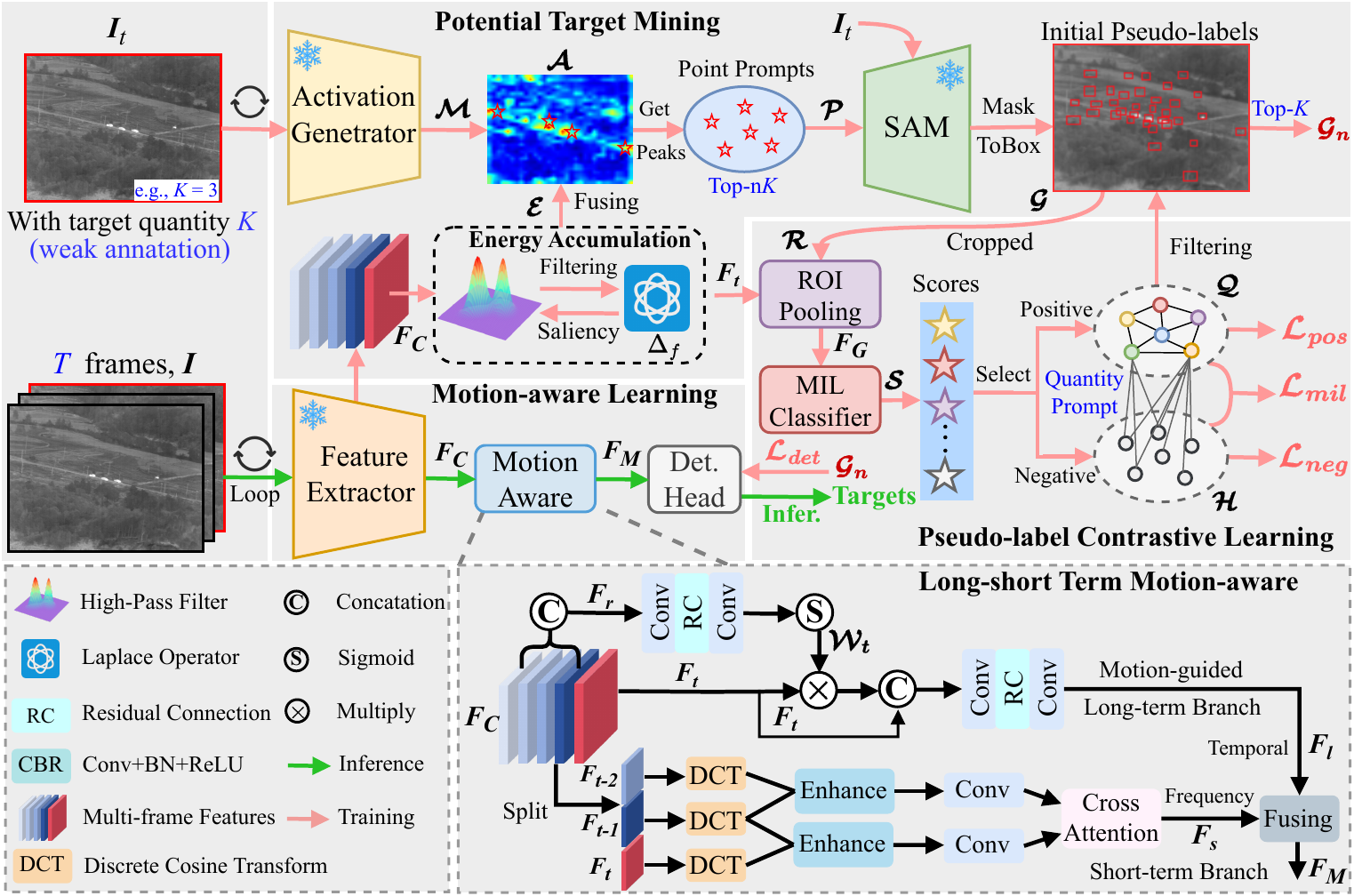}
    \caption{Our WeCoL framework, with training and inference pipelines. 
    In training, the final pseudo-labels $\boldsymbol{\mathcal{G}_n}$, used to supervise multi-frame detector training, are generated and refined by the collaboration of both ``Potential Target Mining'' and ``Pseudo-label Contrastive Learning''.
    In inference, $T$ frames are utilized to detect moving infrared small targets.
    }
    \label{fig:frame}
\end{figure*}
\subsection{Segment Anything Model}
The SAM \cite{SAM} is a prompt-driven large-scale vision foundation model. It could use points, bounding boxes, or texts to guide the model in segmenting any objects within an image.
The core components of SAM contain an image encoder, a lightweight mask decoder, and a flexible prompt encoder.
It has exhibited outstanding performance on a wide range of visual tasks. 
For example, TSP-SAM \cite{TSP-SAM} proposes temporal-spatial SAM for video camouflaged object detection by motion-driven self-prompt learning. Crowd-SAM \cite{crowd-sam} uses SAM as a smart annotator for object detection in crowded scenes.
Moreover, SAM has also been applied to the fields of remote sensing \cite{zhang2024pmho, ma2024sam} and medical images \cite{2024samtest, 2024selfsam}.
Given the good research foundation and strong zero-shot learning ability of SAM, it is strongly anticipated that SAM could be applied to weakly-supervised moving infrared small target detection.

\section{Methodology}
\subsection{Overall Architecture}
Suppose $\boldsymbol{I} = \{\boldsymbol{I}_1, \boldsymbol{I}_2, \cdots, \boldsymbol{I}_t\}$ is a collection of consecutive frames with the time window size of $T$. It is randomly sampled from an infrared video. 
$\boldsymbol{I}_t$ is the keyframe that needs to be detected, and the other ones are its adjacent frames that could provide temporal information. 
Different from previous fully-supervised works, we only have the target quantity in $\boldsymbol{I}_t$ instead of concrete target-wise annotations during model training.

Our primary motivation is to address the challenge of expensive annotation costs in MISTD by using weakly-supervised annotations, $i.e.$, target quantity, to guide model learning.
Specifically, we propose a weakly-supervised contrastive learning framework for moving infrared small targets, $i.e.$, \textbf{WeCoL}, as shown in Figure \ref{fig:frame}.
It follows a general paradigm of WSOD, i.e., enumerate-and-select pseudo-labels \cite{WSDDN, OCIR}. 
These methods often employ traditional region proposal algorithms, e.g., Selective Search \cite{selective}, EdgeBox \cite{edgebox} and MCG \cite{2016MCG}, to enumerate potential proposals and iteratively select high-quality pseudo-labels to promote small target detection performance.

In detail, our WeCoL takes video frames $\boldsymbol{I}$ as inputs. The pretrained CSPDarkNet \cite{ge2021yolox} and InfMAE \cite{liu2024infmae} are used as the visual feature extractor and activation generator, respectively.
Following typical video object detection methods~\cite{transvod}, we extract multi-frame features $\boldsymbol{F_C}\in \mathbb{R}^{T \times C \times H \times W}$ by iteratively feeding each frame into the feature extractor, where $C$, $H$ and $W$ are the channel, height and weight of features, respectively. 
Then, the multi-frame energy response $\boldsymbol{\mathcal{E}}$ is obtained by an energy accumulation unit with the High-Pass Filters (HPF) and Laplace operator. 
Meanwhile, we extract the target activation map $\boldsymbol{\mathcal{M}}$ from a pretrained activation generator ($i.e.$, InfMAE) \cite{liu2024infmae} to provide coarse target locations.
After that, $\boldsymbol{\mathcal{M}}$ and $\boldsymbol{\mathcal{E}}$ are processed in \textbf{\emph{Potential Target Mining} (PTM)} to get peak points $\boldsymbol{\mathcal{P}}$ according to the target quantity in $\boldsymbol{I}_t$ ($i.e.$, $K$). 
Then, SAM utilizes $\boldsymbol{\mathcal{P}}$ as adaptive point prompts to generate the initial pseudo-labels $\boldsymbol{\mathcal{G}}$ of keyframe $\boldsymbol{I}_t$.

Besides, we propose \textbf{\emph{Pseudo-label Contrastive Learning} (PCL)} strategy to select high-quality pseudo-labels as positive samples $\boldsymbol{\mathcal{Q}}$ based on the classification scores $\boldsymbol{\mathcal{S}}$ obtained by a multiple instance learning classifier.
The \emph{Top-K} pseudo-labels corresponding to positive samples are selected as the final ground truth $\boldsymbol{\mathcal{G}_n}$ to supervise detector training.
Then, we calculate the cosine similarity between positive samples as positive contrastive learning loss $\boldsymbol{\mathcal{L}_{pos}}$.
Likewise, the negative contrastive learning loss $\boldsymbol{\mathcal{L}_{neg}}$ is the similarity between positive samples $\boldsymbol{\mathcal{Q}}$ and negative samples $\boldsymbol{\mathcal{H}}$.  

Additionally, we present \textbf{\emph{Long-short Term Motion-aware} (LTM)} to simultaneously model the short-term local motion pattern and long-term global motion trajectory of small targets, enhancing the feature representation ability of detectors. 
The proposed long-short motion fusion strategy could further realize feature complementary, adapting to different complex scenarios. We employ selected pseudo-labels $\boldsymbol{\mathcal{G}_n}$ to supervise the process of motion-aware learning.
Finally, only the trained LTM is used for inference.

\subsection{Potential Target Mining}
How to efficiently obtain pseudo-labels is the key issue of weakly-supervised learning. Previous WSOD methods \cite{WSDDN,OCIR,CBL,PLE} utilize class activation maps or traditional region proposal algorithms \cite{selective,edgebox,2016MCG} to generate proposals, which contain a great number of unreliable pseudo-labels and bring an optimization challenge. 
Moreover, unlike the traditional image-level annotations in WSOD, weakly-supervised MISTD only has the target quantity in each frame as supervision and one target type. 
Therefore, we design the \emph{Potential Target Mining} to explore potential target locations with the help of SAM, as shown in Figure \ref{fig:frame}.

First, the intial activation map $\boldsymbol{\mathcal{M}}$ is obtained by a pretrained infrared foundation model \cite{liu2024infmae}. Then, the multi-frame energy response $\boldsymbol{\mathcal{E}}$ is extracted by an energy accumulation unit. This unit contains a high-pass filter and a Laplace operator. HPF is used to suppress low-frequency background noise and enhance target areas, and the Laplace operator is employed to compute the target saliency of each frame. 
After that, we accumulate the target saliency in multiple time steps, increasing the energy of target regions while gradually smoothing random noise. This calculation process can be formulated as follows:
\begin{equation}
 \left\{\begin{split}
       \boldsymbol{\mathcal{M}} &= f_g(\boldsymbol{I}_t) = Genatrator(\boldsymbol{I}_t) \\
       \boldsymbol{F_H} &= \sum_{i=1}^Tf_h(\boldsymbol{F_C^i}) = HPF(\boldsymbol{F_C}) \\
       \boldsymbol{\mathcal{E}} &= \sum_{i=1}^T \Delta_f(F_H^i(x,y)) =  \sum_{i=1}^T (\frac{\partial^2\boldsymbol{F_H^i}}{\partial x^2}+\frac{\partial^2\boldsymbol{F_H^i}}{\partial y^2})\\
    \end{split} \right.
\end{equation}
where $f_g(\cdot)$ denotes pretrained target activation generators, $f_h(\cdot)$ is high-pass filters, $\boldsymbol{F_H}$ represents the filtered multi-frame features, and $\boldsymbol{\Delta_f}$ is the  Laplace operator.

Second, to avoid excessive false alarms, we integrate the initial activation map $\boldsymbol{\mathcal{M}}$ and multi-frame energy response $\boldsymbol{\mathcal{E}}$ to obtain the final target activation $\boldsymbol{\mathcal{A}}$ with energy accumulating, that is
\begin{equation}
    \label{eq:energy}
    \boldsymbol{\mathcal{A}} = \boldsymbol{\mathcal{M}} + \boldsymbol{\mathcal{E}} \\
\end{equation}
In this way, the designed PTM could simultaneously model both low-level target representations and high-level motion information, ensuring the detected regions of interest (ROI) are significant for locating targets, and remain consistent over time. 

Finally, we could obtain the point prompts $\boldsymbol{\mathcal{P}}$ of SAM by getting the \textit{Top-nK} peaks of $\boldsymbol{\mathcal{A}}$, and then obtain the initial pseudo-labels $\boldsymbol{\mathcal{G}}$, described by following equations:
\begin{equation}
     \left\{\begin{split}
    \boldsymbol{\mathcal{P}} &=\{(x_i,y_i) \mid(x_i,y_i)\in\mathrm{TopK}(\boldsymbol{\mathcal{A}},nK)\} \\
    \boldsymbol{\mathcal{G}} &= \Theta_f(MaskToBox(SAM(\boldsymbol{\mathcal{P}})))
    \end{split} \right.
\end{equation}
where $n \ge 1$ is a constant, $K$ denotes the target quantity in keyframe, TopK(X, Y) means that extracting $Y$ peaks in the feature map $X$, and $\Theta_f$ is a filter used to remove obvious wrong pseudo-labels. 

\subsection{Pseudo-label Contrastive Learning}
In WSOD, the quality of pseudo-labels critically impacts the feature learning process of detectors. Moreover, since infrared small target features are easily confused with backgrounds, training directly with initial pseudo-labels could lead to unreliable feature learning.
As such, we propose \emph{Pseudo-label Contrastive Learning} to further improve the reliability of pseudo-labels and ensure that the positive samples eventually used for model training have high confidence, as shown in Figure \ref{fig:frame}.

First, for each pseudo-label $G_i = (x_l,y_l,x_r,y_r)$, we crop the corresponding region $\mathcal{R}_i$ from the keyframe $\boldsymbol{I}_t$. 
Then, the ROI pooling is performed between cropped regions $\boldsymbol{\mathcal{R}}$ and the keyframe feature $\boldsymbol{F_t}$.
It could be formulated as follows:
\begin{equation}
    \boldsymbol{F_G} = \sum_{i=1}^{m} ROI((Crop(G_i, \boldsymbol{I}_t))) = \sum_{i=1}^{m} ROI((\mathcal{R}_i)),
\end{equation}
where $\boldsymbol{F_G}$ is the feature set of all cropped regions after ROI pooling, and $m$ is the total number of pseudo-labels. 
Inspired by previous methods \cite{WSDDN,OCIR}, we proceed to apply a MIL classifier $f_{mil}(\cdot)$ to assess the quality of each pseudo-label. This is crucial for dividing pseudo-label features into positive and negative samples. Specifically, the above process can be formulated by following equations:
\begin{equation}
    \left\{\begin{split}
        \boldsymbol{\mathcal{S}} &= f_{mil}(\boldsymbol{F_G}) = \sigma(\omega \cdot \boldsymbol{F_G} + b) \\
        \boldsymbol{\mathcal{Q}} &= \{\boldsymbol{F^i_G}\mid \mathcal{S}_i \in TopK(\boldsymbol{\mathcal{S}},k) \}\\
        \boldsymbol{\mathcal{H}} &= \boldsymbol{F_G} \backslash \boldsymbol{\mathcal{Q}} = \{\boldsymbol{F^i_G}\mid \boldsymbol{F^i_G} \notin \boldsymbol{\mathcal{Q}}\},
    \end{split} \right.
\end{equation}
where $\omega$ and $b$ are the weights and bias terms of the classifier, respectively. $\sigma$ is the Sigmoid function.
$\boldsymbol{\mathcal{Q}}$ is selected positive samples and $\boldsymbol{\mathcal{H}}$ is negative samples.

Second, we adopt contrastive learning to further improve the quality of pseudo-labels. 
In detail, we employ positive contrastive learning loss $\mathcal{L}_{pos}$ to enhance the feature consistency between positive samples. The negative contrastive learning loss $\mathcal{L}_{neg}$ is used to separate positive and negative samples in feature subspaces. The calculation process can be formulated as follows:
\begin{equation}
     \left\{\begin{split}
        \mathcal{L}_{pos} &= \sum^K_{i,j\in\boldsymbol{\mathcal{Q}},i\neq j}f_{cos}(\mathcal{Q}_i,\mathcal{Q}_j) = \sum^K_{i,j\in\boldsymbol{\mathcal{Q}},i\neq j} \frac{\mathcal{Q}_i \bullet \mathcal{Q}_j}{\|\mathcal{Q}_i\|_2\,\|\mathcal{Q}_j\|_2} \\
        \mathcal{L}_{neg} &= \sum^m_{i\in\boldsymbol{\mathcal{Q}},j\in\boldsymbol{\mathcal{H}}}f_{cos}(\mathcal{Q}_i,\mathcal{H}_j) = \sum^K_{i\in\boldsymbol{\mathcal{Q}}} \sum^{m-K}_{j\in\boldsymbol{\mathcal{H}}} \frac{\mathcal{Q}_i \bullet \mathcal{H}_j}{\|\mathcal{Q}_i\|_2\,\|\mathcal{H}_j\|_2},
     \end{split} \right.
\end{equation}
where $f_{cos}(\cdot)$ denotes calculating cosine similarity, ``$\bullet$" means inner product and $\| \cdot \|_2$ represents $l_2$ norm.

Moreover, we present the smooth \emph{Top-K} MIL loss $\mathcal{L}_{mil}$ to suppress noise pseudo-labels, enhancing reliable pseudo-label learning from the perspective of target quantity, as follows:
\begin{equation}
    \mathcal{L}_{mil} =-\frac{1}{B}\sum_{b=1}^B\frac{1}{K_b}\sum_{i=1}^{K_b}\log(\omega_{i}+\epsilon), \omega_{i} = Softmax(\mathcal{S}_i),
\end{equation}
where $B$ is the training batch size, $K_b$ is the number of targets ($i.e.$, quantity prompts) in the frame $b$ of current batch, $\omega_{i}$ is the weight coefficients of the $i^{th}$ pseudo-label score ($i.e.$, $\mathcal{S}_i$), and $\epsilon$ is a bias term to ensure numerical stability. 

Finally, the total loss of our PCL could be denoted by
$\mathcal{L}_{pcl} = \mathcal{L}_{pos} + \mathcal{L}_{neg} + \mathcal{L}_{mil}$.
By minimizing $\mathcal{L}_{pcl}$, the model is guided to distinguish targets from backgrounds by contrastive learning. Besides, the selected \emph{Top-K} pseudo-labels $\boldsymbol{\mathcal{G}_n}$ could be used as ground truth to supervise detector training.

\subsection{Long-short Term Motion-aware}
Due to the lack of sufficient annotations in weakly supervised MISTD, only the target quantity is used for detector training. Thus, a key challenge is how to exploit the motion patterns of moving targets sufficiently.
To address this, we propose a \emph{Long-short Term motion-aware} mechanism to simultaneously capture both long and short-term motion cues, as shown in Figure \ref{fig:frame}. 

First, we model the global motion trajectory of moving targets by long-term motion-aware modeling. It could improve the motion representation of targets through adaptively fusing consecutive frames to enhance the keyframe features, as follows:
\begin{equation}
        \left\{\begin{split}
         \boldsymbol{\mathcal{W}_t} &= \sigma (f_r(\boldsymbol{F_r})) = \sigma (f_r(f_c(\boldsymbol{F_1}, \cdots, \boldsymbol{F_{t-1}})))\\
         \boldsymbol{F_l} &= f_r(f_c(\boldsymbol{\mathcal{W}_t} \odot \boldsymbol{F_t}, \boldsymbol{F_t})),
         \end{split} \right.       
\end{equation}
where $\boldsymbol{\mathcal{W}_t}$ is the adaptive weights to enhance keyframe, $f_r(\cdot)$ denotes a residual connection, $f_c(\cdot)$ represents the channel-wise concatenation, ``$\odot$'' is an element-wise product, and $\boldsymbol{F_l}$ means the long-term motion features of moving targets in the temporal domain.
In this manner, we could dynamically adjust the weights of neighboring frames to suppress background noise.  

Second, the short-term motion enhancement based on the frequency domain is proposed to capture the local motion patterns of targets. 
It could extract the frequency components of target features by the discrete cosine transform (DCT) to robustly model local motion patterns, as follows:
\begin{equation}
    \boldsymbol{F'_{t-y}(x)} = \boldsymbol{\mathcal{D}(F_{t-y}(x))} =  \sum_{i=0}^{N-1}\boldsymbol{F_{t-y}(x)}cos(\frac{\pi x}{2N}(2i+1)),
\end{equation}
where $y = \{0,1,2\}$ can control the time window size of short-term motion-aware, $\mathcal{D}(\cdot)$ denotes the DCT, and $N$ is the number of elements in the feature matrix $\boldsymbol{F_{t-y}}$.
Then, the transformed frequency features are further improved in a well-designed enhancement module with high-low frequency decomposing and cross-frame frequency correlation modeling. In detail, taking $\boldsymbol{F_{t-1}}$ and $\boldsymbol{F_{t}}$ as an example, the calculation process is as follows:
\begin{equation}
    \left\{\begin{split}
    \boldsymbol{F_{t-1}^l},\boldsymbol{F_{t-1}^h}&,\boldsymbol{F_{t}^l},\boldsymbol{F_{t}^h} = Divide (\boldsymbol{F_{t-1}}, \boldsymbol{F_{t}}) \\
    \boldsymbol{F_{(t-1) \leftrightarrow t}^l} &= f_r(MLP(Atten(\boldsymbol{F_{t-1}^l},\boldsymbol{F_{t}^l})))\\
    \boldsymbol{F_{(t-1) \leftrightarrow t}^h} &= f_r(MLP(Atten(\boldsymbol{F_{t-1}^h},\boldsymbol{F_{t}^h})))\\
    \boldsymbol{F_{(t-1) \leftrightarrow t}} &= f_r(f_c(\boldsymbol{F_{(t-1) \leftrightarrow t}^l}, \boldsymbol{F_{(t-1) \leftrightarrow t}^h})),
    \end{split} \right. 
\end{equation}
where $Divide(\cdot)$ denotes dividing features into high and low frequency groups in channel dimensions, $MLP(\cdot)$ is a linear layer, and $Atten(\cdot)$ represents the multi-head self-attention.
Similarly, we could process $\boldsymbol{F_{t-2}}$ and $\boldsymbol{F_{t-1}}$ to obtain $\boldsymbol{F_{(t-2) \leftrightarrow (t-1)}}$. After that, the short-term motion features $\boldsymbol{F_s}$ are obtained by the cross attention, as follows:
\begin{equation}
    \boldsymbol{F_s} = f_*(CrossAtt(f_*(\boldsymbol{F_{(t-2) \leftrightarrow (t-1)}}),f_*(\boldsymbol{F_{(t-1) \leftrightarrow t}}))),
\end{equation}
where $CrossAtt(\cdot)$ is the cross attention to mine the semantic associations between two features, and $f_*(\cdot)$ denotes convolutions.

Finally, we utilize the long-short motion fusion strategy to combine the local motion patterns and global motion trajectory effectively. The calculation process can be denoted as follows:
\begin{equation}
    \boldsymbol{F_M} = Fusing(\boldsymbol{F_l},\boldsymbol{F_s}) =SA(CA(f_*(f_c(\boldsymbol{F_l},\boldsymbol{F_s})))),
\end{equation}
where $\boldsymbol{F_M}$ denotes the comprehensive motion features, $SA(\cdot)$ is spatial attention, and $CA(\cdot)$ represents channel attention.
\begin{table*}[h]
\centering
\caption{The details of DAUB and ITSDT-15K datasets.}
\label{tab:dataset}
\resizebox{0.95\linewidth}{!}{
\begin{tabular}{c|c|c|c|c|c|c|c}
\toprule
\textbf{Dataset} &
  \textbf{Annotations} &
  \textbf{Sequential} &
  \textbf{Classes} &
  \textbf{Target size} &
  \textbf{Target type} &
  \textbf{Number of frames} &
  \textbf{Scene description} \\ \midrule
  DAUB &
  target quantity &
  \textcolor{black}{\ding{51}} &
  1 &
  1 $\sim$ 10 pixels &
  drone &
  13,778 &
  \begin{tabular}[c]{@{}c@{}}ground background, buildings, \\ air background, plain, suburb, \\ air ground junction background\end{tabular} \\ \midrule
ITSDT-15K &
  target quantity &
  \textcolor{black}{\ding{51}} &
  1 &
  1 $\sim$ 16 pixels &
  vehicle &
  15,000 &
  \begin{tabular}[c]{@{}c@{}}ground background, forest, \\ plain, suburb, traffic \\ air ground junction background\end{tabular} \\ 
\bottomrule
\end{tabular}}
\end{table*}

\subsection{Loss Function}
We use the selected \emph{Top-K} high-quality pseudo-labels $\boldsymbol{\mathcal{G}_n}$ to supervise detector training. The total training loss of our proposed WeCoL could be described as follows:
\begin{equation}
    \mathcal{L} = \eta\mathcal{L}_{det} + \gamma\mathcal{L}_{pcl} 
    = \eta\mathcal{L}_{det} + \gamma(\mathcal{L}_{pos} + \mathcal{L}_{neg} + \mathcal{L}_{mil}),
\end{equation}
where $\eta$ and $\gamma$ are two hyper-parameters to balance loss terms, and $\mathcal{L}_{det}$ is the detection loss based on YOLOX \cite{ge2021yolox}, As such, $\mathcal{L}_{det}$ could be denoted as follows:
\begin{equation}
\label{eq:detec}
    \mathcal{L}_{det} = \mathcal{L}_{cls}+ \lambda_1\mathcal{L}_{reg} + \lambda_2\mathcal{L}_{obj},
\end{equation}
where $\lambda_1$ and $\lambda_2$ are two weight coefficients, $\mathcal{L}_{cls}$ is a classification loss, $\mathcal{L}_{reg}$ is a bounding box regression loss, and $\mathcal{L}_{obj}$ is a target probability loss.
Following the default setting of YOLOX, we employ sigmoid focal loss \cite{lin2017focalloss} for $\mathcal{L}_{cls}$ and $\mathcal{L}_{obj}$, and GIoU \cite{GIOU} as the regression loss $\mathcal{L}_{reg}$.

\section{Experiments}
\subsection{Datasets and Evaluation Metrics}
We evaluate our WeCoL on two public MISTD datasets: DAUB \cite{daub} and ITSDT-15K \cite{ITSDT}. 
The DAUB dataset has a training set of 10 videos with 8,983 frames and a test set of 7 videos with 4,795 frames.
The ITSDT-15K dataset contains a training set of 40 videos with 10,000 frames and a test set of 20 videos with 5,000 frames. The details of three datasets are shown in Table \ref{tab:dataset}.

For a fair comparison with other detectors, following previous works \cite{tridos,sttrans}, we use the standard evaluation metrics in object detection paradigm, $i.e.$, Precision ($Pr$), Recall ($Re$), F1 score and $\text{mAP}_{50}$ (the mean Average Precision with an IoU threshold 0.5).
These metrics can be denoted as follows:
\begin{align}
\begin{split}
    \text{Precision} &= \frac{\text{TP}}{\text{TP}+\text{FP}}\\
\text{Recall} &=  \frac{\text{TP}}{\text{TP}+\text{FN}}\\
\text{F1} &= \frac{2\times \text{Precision} \times \text{Recall}}{\text{Precision}+\text{Recall}}
\end{split}
\end{align}
where TP, FP and FN represent the number of correct detection targets (True positive), false alarms (False positive) and missed detection targets (False negative), respectively. F1 score is a comprehensive metric, combining both precision and recall.
\subsection{Implementation Details}

For all compared methods, the resolution of input frames is resized to $512 \times 512$.  We modify segmentation-based methods into a detection paradigm by adding a detection head to generate bounding boxes.
Moreover, SGD is adopted as the optimizer with an initial learning rate of 0.01, a momentum of 0.937, and a weight decay of $5 \times 10^{-4}$.
We train our WeCoL for 100 epochs with a batch size of 4. The model weights are initialized with the usual
distribution.
For hyper-parameters, the time window size $T$ of frame sampling is 5. Following the default setting of detection head \cite{ge2021yolox}, $\lambda_1$ and $\lambda_2$ in Eq. (\ref{eq:detec}) are 5 and 1, respectively. $n$, $\eta$ and $\gamma$ are set to 3, 1 and 1, respectively.
Regarding hardware, we conduct all experiments on two Nvidia GeForce 4090D GPUs.

\subsection{Comparisons With Other Methods}
\begin{table*}[h]
\centering
\caption{Quantitative detection performance comparisons. The best and second-best results are highlighted in bold and underlined, respectively. All compared weakly-supervised methods only use image-level annotations.
}
\label{tab:quantitave}
\resizebox{0.95\linewidth}{!}{
\begin{tabular}{c|ll|cccc|cccc}
\toprule
\multirow{2}{*}{\textbf{Scheme}} &
  \multirow{2}{*}{\textbf{Methods}} &
  \multirow{2}{*}{\textbf{Publication}} &
  \multicolumn{4}{c}{\textbf{DAUB}} &
  \multicolumn{4}{c}{\textbf{ITSDT-15K}} \\
 &
   &
   &
  $\textbf{mAP}_{\textbf{50}}$ &
  \textbf{Precision} &
  \textbf{Recall} &
  \textbf{F1} &
  $\textbf{mAP}_{\textbf{50}}$ &
  \textbf{Precision} &
  \textbf{Recall} &
  \textbf{F1} \\ \midrule
\multirow{6}{*}{Model-driven}      
& MaxMean \cite{MaxMean} & SPIE 1999       & 10.71 & 20.38 & 53.87 & 29.57 & 0.87  & 10.85 & 8.74  & 9.68  \\ 
                                   & TopHat \cite{Tophat}  & IPT 2006        & 16.99 & 21.69 & 79.83 & 34.11 & 11.61 & 27.21 & 43.07 & 33.35 \\
                                   & RLCM \cite{han2018infraredrlcm}    & IEEE TGRS 2013  & 0.02  & 0.27  & 5.21  & 0.51  & 4.62  & 15.38 & 30.76 & 20.50 \\
                                   & HBMLCM \cite{HBMLCM}  & IEEE GRSL 2019  & 3.90  & 23.96 & 16.52 & 19.56 & 0.72  & 7.97  & 9.37  & 8.61  \\
                                   & PSTNN \cite{PSTNN}   & RS 2019         & 17.31 & 25.56 & 68.86 & 37.28 & 7.99  & 22.98 & 35.21 & 27.81 \\
                                   & WSLCM \cite{wslcm}   & SP 2020         & 1.37  & 11.88 & 11.57 & 11.73 & 2.36  & 16.78 & 14.53 & 15.58 \\ \midrule
\multirow{14}{*}{Fully-supervised} & ACM \cite{ACM}     & WACV 2021       & 64.02 & 70.96 & 91.30 & 79.86 & 55.38 & 78.37 & 71.69 & 74.88 \\
                                   & RISTD \cite{hou2021ristdnet}   & IEEE GRSL 2022  & 81.05 & 83.46 & \underline{98.27} & 90.26 & 60.47 & 85.49 & 71.60 & 77.93 \\
                                   & ISTDUNet \cite{ISTDU}   & IEEE GRSL 2022  & 82.47 & 90.91 & 91.57 & 91.24 & 64.74 & 82.73 & 80.02 & 81.35 \\
                                   & ISNet \cite{ISNetzhang2022isnet}   & CVPR 2022       & 83.43 & 89.36 & 94.99 & 92.09 & 62.29 & 83.46 & 75.32 & 79.18 \\
                                   & UIUNet \cite{wu2022uiuNet}  & IEEE TIP 2022   & 86.41 & 94.46 & 92.03 & 93.23 & 65.15 & 84.07 & 78.39 & 81.13 \\
                                   & SANet \cite{zhu2023SAnet}   & ICASSP 2023     & 87.12 & 93.44 & 94.93 & 94.18 & 62.17 & 87.78 & 71.23 & 78.64 \\
                                   & AGPCNet \cite{AGPCNet} & IEEE TAES 2023  & 76.72 & 82.29 & 94.43 & 87.95 & 67.27 & \underline{91.19} & 74.77 & 82.16 \\
                                   & RDIAN \cite{RDIAN}   & IEEE TGRS 2023  & 84.92 & 88.20 & 97.27 & 92.51 & 68.49 & 90.56 & 76.06 & 82.68 \\
                                   & DNANet \cite{dnanet}  & IEEE TIP 2023   & 89.93 & 92.49 & \underline{98.27} & 95.29 & 70.46 & 88.55 & 80.73 & 84.46 \\
                                    & CSVIG \cite{lin2024csvig} &  ESWA 2024 &  79.98 &  84.86 &  95.14 &  89.71 &
  60.11 &  76.92 &  74.98 &  75.89  \\
    &  SCTrans \cite{sctransnet} &  IEEE TGRS 2024 &  83.27 &  92.04 &  91.72 &  91.88 &
  71.37 &  \textbf{91.74} &  78.49 &  84.60 \\
  &  SSTNet \cite{sstnet}  &  IEEE TGRS 2024 &  \textbf{95.59} &  \textbf{98.08} &  98.10 &  \textbf{98.09} &
  \textbf{76.96} &  91.05 &  85.29 &  \textbf{88.07}\\
                                   & SIRST5K \cite{lu2024sirst5k} & IEEE TGRS 2024  & 93.31 & \underline{97.78} & 96.93 & 97.35 & 61.52 & 86.95 & 71.32 & 78.36 \\
                                   & MSHNet \cite{MSHNet}  & CVPR 2024       & 85.97 & 93.13 & 93.12 & 93.13 & 60.82 & 89.69 & 68.44 & 77.64 \\
                                   & RPCANet \cite{rpcanet} & WACV 2024       & 85.98 & 89.38 & 97.56 & 93.29 & 62.28 & 81.46 & 77.10 & 79.22 \\
                                   &  ST-Trans \cite{sttrans}  &  IEEE TGRS 2024 &  92.73 &  97.75 &  95.52 &  96.62 &
  \underline{76.02} &  89.96 &  85.18 &  \underline{87.50}\\
                                   & DTUM \cite{DTUM}    & IEEE TNNLS 2025 & 85.86 & 87.54 & \textbf{99.79} & 93.26 & 67.97 & 77.95 & \textbf{88.28} & 82.79 \\
                                   & MLPNet \cite{wang2024mlp}  & IEEE TGRS 2025  & \underline{93.58} & 97.08 & 97.89 & \underline{97.49} & 53.76 & 74.06 & 73.19 & 73.63 \\
                                   & LSKNet \cite{wu2024lsk}  & IEEE TGRS 2025  & 84.66 & 88.84 & 96.04 & 92.30 & 66.07 & 90.75 & 73.72 & 81.35 \\ \midrule
\multirow{5}{*}{Weakly-supervised} & \emph{SPE} \cite{liao2022spe}     & \emph{ECCV 2022}       & -     & -     & -     & -     & -     & -     & -     & -     \\
                                   & \emph{CBL} \cite{CBL}     & \emph{ICCV 2023}       & -     & -     & -     & -     & -     & -     & -     & -     \\
                                   & ESFL \cite{ESFL}    & IEEE TNNLS 2024 & 9.87  & 18.65 & 26.72 & 21.97 & 3.48  & 12.73 & 17.52 & 14.75 \\
                                   & PLE \cite{PLE}     & KBS 2025        & 29.46 & 43.76 & 53.64 & 48.20 & 22.31 & 34.72 & 38.56 & 36.54 \\
                                   \rowcolor{blue!10}
                                   & \textbf{WeCoL} (\textbf{Ours})    & -               & 89.41 & 92.13 & 98.25 & 95.09 & 71.28 & 82.74 & \underline{87.69} & 85.14
                                   \\ \bottomrule
\end{tabular}}
\end{table*}
\subsubsection{Quantitative Comparison}
As shown in Table \ref{tab:quantitave}, we present the quantitative comparisons between our proposed WeCoL and some top-performing methods, covering model-driven, fully-supervised, and weakly-supervised ones.
Importantly, all weakly-supervised methods only use image-level annotations for detector training.  
From table, we could have two obvious observations.
One is that our WeCoL significantly surpasses all other weakly supervised methods. It could even outperform many early fully supervised methods and achieve over 90\% SOTA performance across most metrics on two datasets.
For example, on DAUB, our WeCoL achieves $\text{mAP}_{50}$ 89.41\% and F1 95.09\%, nearly 94\% of SOTA fully-supervised methods, $i.e.$, $\text{mAP}_{50}$ 95.59\% and F1 98.09\% by SSTNet \cite{sstnet}.
In terms of $Re$, the 98.25\% by WeCoL is slightly lower than the SOTA 99.79\% by DTUM \cite{DTUM}.

The other is that current general weakly-supervised learning methods become low effective in the highly challenging scenarios of MISTD.
For instance, due to some weakly-supervised schemes primarily focusing on static objects, they cannot even obtain specific numerical results on DAUB and ITSDT-15K, such as SPE \cite{liao2022spe} and CBL \cite{CBL}. 
Moreover, ESFL \cite{ESFL} only obtains $\text{mAP}_{50}$ 9.87\% and F1 21.97\% on DAUB. 
On challenging ITSDT-15K, the 
$\text{mAP}_{50}$ and F1 of PLE \cite{PLE} are 22.31\% and 36.54\%, respectively. 
In contrast, our WeCoL could achieve an F1 of 85.14\%, only 2.93\% lower than that by the SOTA fully-supervised methods, $i.e.$, SSTNet \cite{sstnet}.

\begin{table}[h]
\centering
\caption{Inference cost comparisons on ITSDT-15K.}
\label{tab:complex}
\resizebox{\linewidth}{!}{
\begin{tabular}{l|c|cc|ccc}
\toprule
\textbf{Methods} & \textbf{Frames} & $\textbf{mAP}_{\textbf{50}}$ $\uparrow$ & \textbf{F1} $\uparrow$ & \textbf{Params} $\downarrow$ & \textbf{GFlops} $\downarrow$ & \textbf{FPS} $\uparrow$ \\ \midrule
ACM \cite{ACM}               & 1 & 55.38          & 74.88          & \underline{3.04M}          & \textbf{24.73} & \underline{29.11}          \\
                                  RISTD \cite{hou2021ristdnet}           & 1 & 60.47          & 77.93          & 3.28M          & 76.28          & 10.21          \\
                                   ISNet \cite{ISNetzhang2022isnet}             & 1 & 62.29          & 79.18          & 3.49M          & 265.73         & 11.20          \\
                                   UIUNet \cite{wu2022uiuNet}           & 1 & 65.15          & 81.13          & 53.06M         & 456.70         & 3.63           \\
                                   ISTDUNet \cite{ISTDU}  & 1 & 64.74          & 81.35          & 5.27M         & 394.32         & 7.01 \\
                                  SANet \cite{zhu2023SAnet}            & 1 & 62.17          & 78.64          & 12.40M         & 42.04          & 10.55          \\
                                   AGPCNet \cite{AGPCNet}        & 1 & 67.27          & 82.16          & 14.88M         & 366.15         & 4.79           \\
                                   RDIAN \cite{RDIAN}             & 1 & 68.49          & 82.68          & \textbf{2.74M} & 50.44          & 20.52          \\
                                   DNANet \cite{dnanet}             & 1 & 70.46          & 84.46          & 7.22M          & 135.24         & 4.82           \\
                                   CSVIG \cite{lin2024csvig}            & 1 & 60.11          & 75.89          & 5.81M          & 117.56         & 12.70          \\
                                   SIRST5K \cite{lu2024sirst5k}           & 1 & 61.52          & 78.36          & 11.48M         & 182.61         & 7.37           \\
                                   MSHNet \cite{MSHNet}           & 1 & 60.82          & 77.64          & 6.59M          & 69.59          & 18.55          \\
                                   RPCANet \cite{rpcanet}          & 1 & 62.28          & 79.22          & 3.21M          & 382.69         & 15.89          \\
                                   MLPNet \cite{wang2024mlp}          & 1 & 53.76          & 73.63          & 10.79M          & \underline{34.72}         & 5.93          \\
                                   LSKNet \cite{wu2024lsk}          & 1 & 66.07         & 81.35          & 3.42M          & 76.00         & \textbf{40.63}          \\
                                   DTUM \cite{DTUM}           & 5 & 67.97          & 82.79          & 9.64M          & 128.16         & 14.28          \\
                                    SSTNet \cite{sstnet}           & 5 & \textbf{76.96}          & \textbf{88.07}          & 11.95M          & 123.60         & 7.37          \\
                                   ST-Trans \cite{sttrans}          & 5 & \underline{76.02}          & \underline{87.50}          & 38.13M         & 145.16         & 3.90           \\ \midrule
ESFL \cite{ESFL}          & 1 & 3.48          & 14.75          & 32.04M         & 63.65         & 9.56\\
PLE \cite{PLE}     & 1 & 22.31          & 36.54          & 42.75M         & 75.78         & 10.25 \\
\rowcolor{blue!10}
\textbf{WeCoL (Ours)} & 5 & 71.28          & 85.14          & 11.16M         & 99.34         & 16.92\\
 \bottomrule      
\end{tabular}}
\end{table}

\begin{figure}[h]
    \centering
\includegraphics[width=0.93\linewidth]{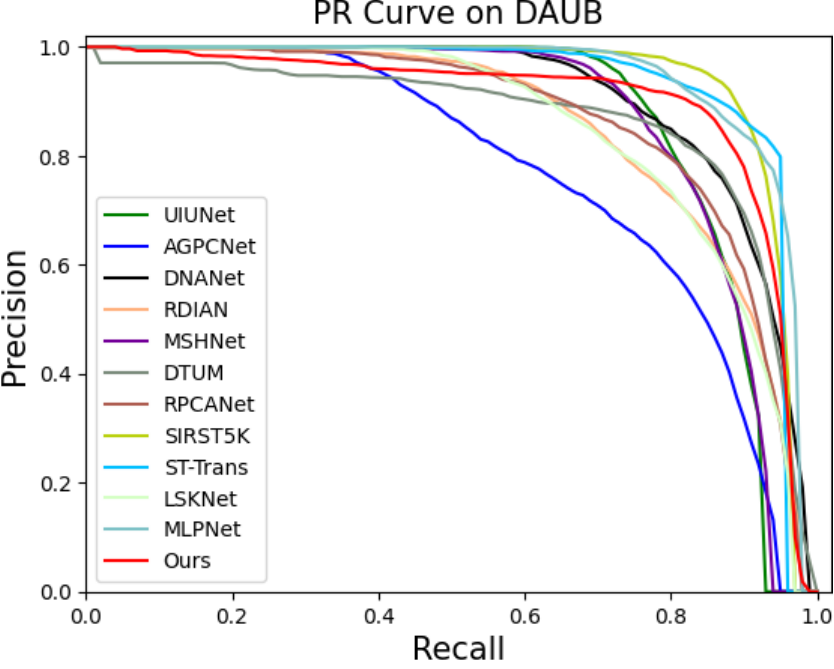}
    \caption{The PR curves comparisons for different methods on DAUB.}
    \label{fig:pr1}
\end{figure}
\begin{figure}[h]
    \centering
\includegraphics[width=0.93\linewidth]{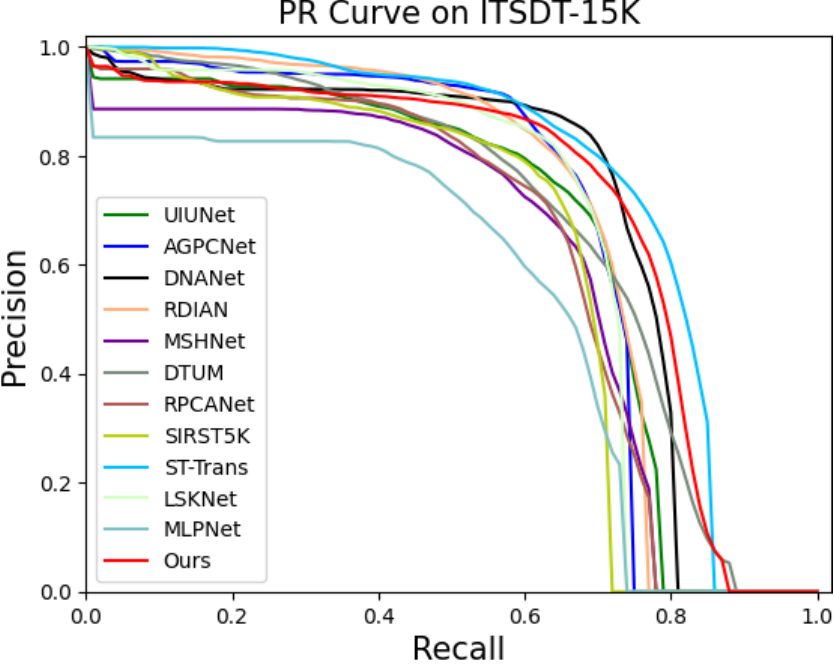}
    \caption{The PR curves comparisons for different methods on ITSDT-15K.}
    \label{fig:pr2}
\end{figure}
\begin{table*}[h]
\centering
\caption{The ablation study on our WeCoL. 
\textbf{Based on SAM,} $P_m$ denotes initial activation maps $\boldsymbol{\mathcal{M}}$, and $P_e$ represents multi-frame energy accumulation $\boldsymbol{\mathcal{E}}$. 
$C_p$ is positive contrastive learning, $C_n$ is negative contrastive learning and $C_m$ represents smooth TopK MIL.
$M_s$ is short-term motion capturing, and $M_l$ denotes long-term motion modeling.
}
\label{tab:ablation}
\resizebox{0.97\linewidth}{!}{
\begin{tabular}{l|cc|ccc|cc|cccccccc}
\toprule
\multirow{2}{*}{\textbf{Settings}} &
  \multicolumn{2}{c|}{\textbf{PTM}} &
  \multicolumn{3}{c|}{\textbf{PCL}} &
  \multicolumn{2}{c|}{\textbf{LTM}} &
  \multicolumn{4}{c}{\textbf{DAUB}} &
  \multicolumn{4}{c}{\textbf{ITSDT-15K}} \\ \cmidrule{2-16}
 &
  $\boldsymbol{P_m}$ &
  $\boldsymbol{P_e}$ &
  $\boldsymbol{C_p}$ &
  $\boldsymbol{C_n}$ &
  $\boldsymbol{C_m}$ &
  $\boldsymbol{M_s}$ &
  $\boldsymbol{M_l}$ &
  $\textbf{mAP}_{\textbf{50}}$ &
  \textbf{Pr} &
  \textbf{Re} &
  \textbf{F1} &
  $\textbf{mAP}_{\textbf{50}}$ &
  \textbf{Pr} &
  \textbf{Re} &
  \textbf{F1} \\ \midrule
\textbf{w/o All} &
  - &
  - &
  - &
  - &
  - &
  - &
  - &
  0.00 &
  0.00 &
  0.00 &
  0.00 &
  0.00 &
  0.00 &
  0.00 &
  0.00 \\ \midrule
\multirow{2}{*}{\textbf{w P}} &
  \checkmark &
  - &
  - &
  - &
  - &
  - &
  - &
  63.24 &
  82.93 &
  77.25 &
  79.79 &
  52.38 &
  74.89 &
  71.57 &
  73.19 \\ 
 &
  \checkmark &
  \checkmark &
  - &
  - &
  - &
  - &
  - &
  67.19 &
  83.75 &
  81.96 &
  82.85 &
  54.76 &
  75.70 &
  73.34 &
  74.50 \\ \midrule
\multirow{3}{*}{\textbf{w P \& C}} &
  \checkmark &
  \checkmark &
  \checkmark &
  - &
  - &
  - &
  - &
  70.33 &
  83.41 &
  85.94 &
  84.66 &
  59.44 &
  75.63 &
  79.98 &
  77.74 \\
 &
  \checkmark &
  \checkmark &
  \checkmark &
  \checkmark &
  - &
  - &
  - &
  78.49 &
  88.68 &
  88.82 &
  88.75 &
  64.05 &
  75.52 &
  86.24 &
  80.52 \\
 &
  \checkmark &
  \checkmark &
  \checkmark &
  \checkmark &
  \checkmark &
  - &
  - &
  84.95 &
  87.67 &
  97.78 &
  92.45 &
  66.07 &
  79.72 &
  83.35 &
  81.49 \\ \midrule
\multirow{2}{*}{\textbf{w P \& C \& M}} &
  \checkmark &
  \checkmark &
  \checkmark &
  \checkmark &
  \checkmark &
  \checkmark &
  - &
  86.73 &
  89.84 &
  97.65 &
  93.58 &
  68.87 &
  \textbf{84.41} &
  82.33 &
  83.36 \\
 &
  \checkmark &
  \checkmark &
  \checkmark &
  \checkmark &
  \checkmark &
  - &
  \checkmark &
  87.85 &
  91.78 &
  96.96 &
  94.29 &
  69.09 &
  81.83 &
  85.01 &
  83.39 \\ \midrule
   \rowcolor{blue!10}
\textbf{w All} &
  \checkmark &
  \checkmark &
  \checkmark &
  \checkmark &
  \checkmark &
  \checkmark &
  \checkmark &
  \textbf{89.41} &
  \textbf{92.13} &
  \textbf{98.25} &
  \textbf{95.09} &
  \textbf{71.28} &
  82.74 &
  \textbf{87.69} &
  \textbf{85.14} \\ \bottomrule
\end{tabular}}
\end{table*}

\subsubsection{Inference Cost Comparison}
The inference cost comparisons on 15 representative methods are presented in Table \ref{tab:complex}, revealing two obvious findings.
One is that although our WeCoL utilizes a sequence of five frames, the number of parameters remains moderate.
For example, our model has 11.16M parameters, higher than the SOTA method ACM (3.04M).
However, most compared methods surpass 10M parameters, $e.g.$, UIUNet with 53.06M, ST-Trans with 38.13M, and PLE with 42.75M. Besides, our WeCoL has a middle FPS of 16.92, which is higher than the 5.93 by MLPNet, and the 9.56 by ESFL.
One possible reason is that proposed potential target mining and contrastive learning only work in training phase. We just employ motion-aware learning for inference.

The other is that the use of multi-frame images often results in a high number of FLOPs, while incorporating multi-frame features could improve model performance.
For instance, our WeCoL has 99.34 GFlops, higher than most single-frame methods, such as MSHNet with 69.59 GFlops and LSKNet with 76.00 GFlops. 
However, our $\text{mAP}_{50}$ and F1 are higher than all those methods.
Therefore, these associated inference costs are reasonable and worthwhile given the significant performance gains.

\subsubsection{PR Curves Comparison}
As usual, we employ precision-recall (PR) curves to visually assess the comprehensive performance of various methods, as shown in Figure \ref{fig:pr1} and \ref{fig:pr2}. 
By comparison, we could observe that our curves are above other ones in most situations. 
The closer a method is to the top-right corner, the higher its validity.
Therefore, the group of PR curves indicate that our WeCoL has a comparable performance to SOTA fully-supervised methods, 
a good complexity-performance trade-off is achieved.

\subsubsection{Visual Comparison}
\begin{figure*}[t]
    \centering
    \includegraphics[width=0.98\linewidth]{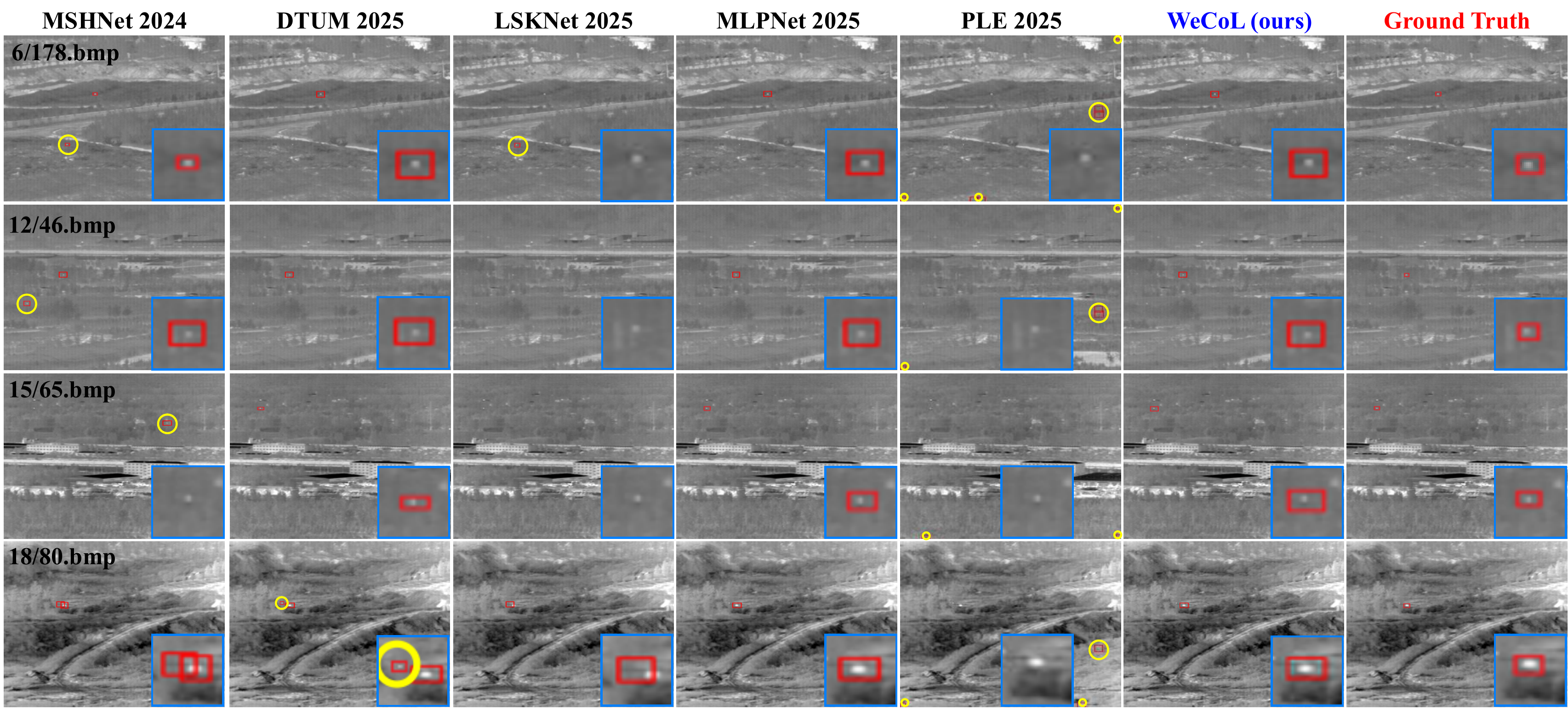}
    \caption{The visual results of different methods on DAUB. Blue boxes and yellow circles represent amplified target regions and false alarms, respectively.
    }
    \label{fig:vis1} 
\end{figure*}

\begin{figure*}[t]
    \centering
    \includegraphics[width=0.98\linewidth]{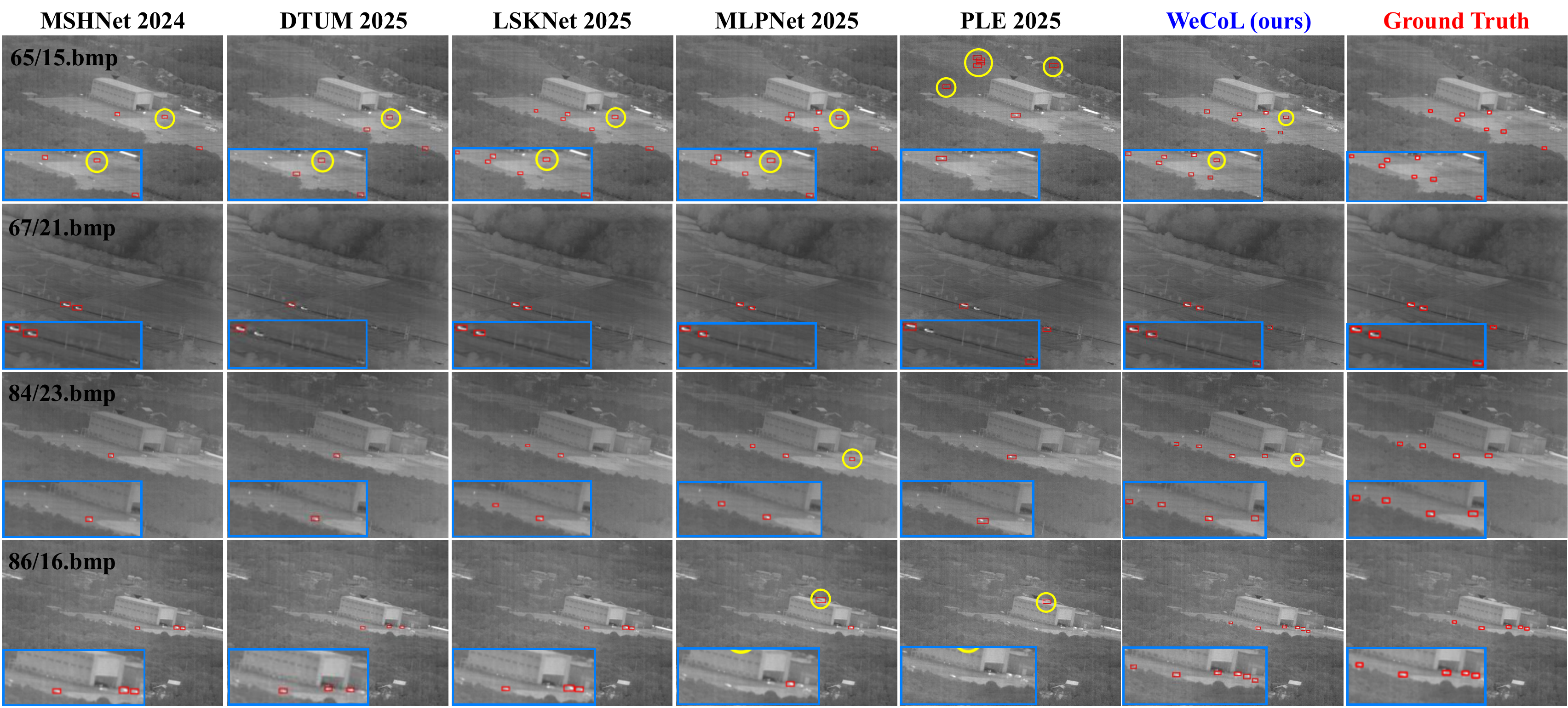}
    \caption{The visual results of different methods on ITSDT-15K. Blue boxes and yellow circles represent amplified target regions and false alarms, respectively.
    }
    \label{fig:vis2} 
\end{figure*}

For intuitiveness, we select 5 representative methods to visually compare with our WeCoL on two datasets, as depicted in Figure \ref{fig:vis1} and \ref{fig:vis2}.
We choose several typical scenarios from DAUB and ITSDT-15K, including mountains, plains and suburbs. 

From these two figures, it is evident that our method could usually precisely detect moving small targets at the same IoU settings, while others could often produce missed detections and false alarms. 
For example, in Figure \ref{fig:vis1}, on the sample  6/178.bmp of DAUB, our WeCoL could detect the correct one target, while MSHNet \cite{MSHNet} detects two targets, and PLE \cite{PLE} even detects 5 targets, causing false detections. 
Besides, LSKNet \cite{wu2024lsk} mistakenly treats a bright spot as a target.
Moreover, in Figure \ref{fig:vis2}, on challenging ITSDT-15K, our WeCoL could still correctly detect almost all targets in different cases. However, MLPNet \cite{wang2024mlp} and DTUM \cite{DTUM} appear miss detections, $e.g.$, on the sample 65/15.bmp of ITSDT-15K. 
Furthermore, it could be clearly seen that the results of our method are highly consistent with the ground truth.
In summary, these visualization comparisons above consistently support the quantitative results in Table \ref{tab:quantitave}. It further confirms the effectiveness of our WeCoL compared to other methods in detecting moving infrared small targets.


\subsection{Ablation Study}
\subsubsection{Effects of Different Components}
To analyze the impacts of different components in our WeCoL, we conduct a series of ablation studies on DAUB and ITSDT-15K, as shown in Table \ref{tab:ablation}.
From it, we could have two obvious findings.
One is that all components are consistently effective in improving detection performance.
For example, the baseline (``w/o All'') has a detection performance 0, indicating that it cannot effectively train.
Then, we progressively add these components to the baseline and observe their impacts.
After integrating \emph{Potential Target Mining} (``w $P_m$ \& $P_e$''), on DAUB, the $\text{mAP}_{50}$ and F1 increase to 67.19\% and 82.85\%, respectively. It demonstrates that our PTM could adequately mine the potential pseudo-labels of small targets.  
The setting (``w P \& C'') further raises $\text{mAP}_{50}$ to 84.95\% and F1 to 92.45\%.
Besides, the proposed \emph{Motion-aware Learning} (``w P \& C \& $\text{M}_l$'') could also have an apparent gain with $\text{mAP}_{50}$ 87.46\% and F1 97.06\%.

The other is that the best performance could be obtained by combining all components.
For instance, on ITSDT-15K, when all components are fully assembled ($i.e.$, ``w All''), detection performance can reach a peak, with $\text{mAP}_{50}$ 71.28\% and F1 score 85.14\%. Similarly, on DAUB, the $\text{mAP}_{50}$ and F1 score could also be refreshed to 89.41\% and 95.09\%, respectively.

\subsubsection{Impacts of Different Targets Mining Methods}
To explore the impacts of different target mining methods and further verify that the performance improvement is due to our innovative approach, rather than the simple use of pretrained large models, $i.e.$, InfMAE and SAM, we conduct a group of experiments, as shown in Table \ref{tab:targets}.

From experimental results, it is obvious that 
the pseudo-labels generated by the ``Selective Search'' could not generate reliable pseudo-labels effectively, nor any numerical results. 
It takes the longest time ($i.e.$, 1912s) and cannot effectively supervise the detector training. One possible reason is that traditional region proposal algorithms are mainly designed for general objects and are not suitable for small targets.
Besides, using energy accumulation $\boldsymbol{\mathcal{E}}$, initial activation map $\boldsymbol{\mathcal{M}}$, final target activation $\boldsymbol{\mathcal{A}}$, or SAM alone is not effective for generating the reliable pseudo-labels of infrared small targets.

For example, on DAUB, using energy accumulation alone just obtains $\text{mAP}_{50}$ 5.32\% and F1 14.48\%. 
Furthermore, only utilizing SAM merely achieves $\text{mAP}_{50}$ 32.13\% and F1 44.06\%. 
In contrast, our PTM could obtain the best $\text{mAP}_{50}$ 89.41\% and F1 95.09\% with medium time cost ($i.e.$, 651s).
On challenging ITSDT-15K, our PTM still achieves a peak with $\text{mAP}_{50}$ 71.28\% and F1 85.14\%. It indicates that our PTM can effectively explore potential targets, rather than relying solely on pre-trained large models.
\begin{table*}[h]
\centering
\caption{Ablation study on different target mining methods. ``Time'' denotes the average time spent in generating pseudo-labels.
}
\label{tab:targets}
\resizebox{0.83\linewidth}{!}{
\begin{tabular}{l|ccccccccc}
\toprule
\multicolumn{1}{l}{\multirow{2}{*}{\textbf{Settings}}} &
  \multicolumn{4}{c}{\textbf{DAUB}} &
  \multicolumn{4}{c}{\textbf{ITSDT-15K}} & \multirow{2}{*}{\textbf{Time}} \\ \cmidrule{2-9}
\multicolumn{1}{c}{} &
  $\textbf{mAP}_{\textbf{50}}$ &
  \textbf{Pr} &
  \textbf{Re} &
  \textbf{F1} &
  $\textbf{mAP}_{\textbf{50}}$ &
  \textbf{Pr} &
  \textbf{Re} &
  \textbf{F1} \\ \midrule
Only w/ Selective Search    & -     & -     & -     & -     & -     & -     & -     & -   & 1912s  \\
Only w/ Energy Accumulation $\boldsymbol{\mathcal{E}}$ 
& 5.32  & 12.37 & 17.45 & 14.48 & 3.45  & 8.95  & 17.68 & 11.88 & \textbf{359s} \\
Only w/ Activation Generator $\boldsymbol{\mathcal{M}}$     
& 8.71  & 14.62 & 19.83 & 16.83 & 4.97  & 13.32 & 22.10 & 16.62 & 558s \\
Only w/ Target Activation $\boldsymbol{\mathcal{A}}$     & 21.89  & 23.37 & 28.53 & 25.69 & 13.83  & 19.74 & 23.48 & 21.45 & 619s \\
Only w/ SAM            & 32.13 & 39.64 & 49.58 & 44.06 & 15.46 & 19.31 & 27.65 & 22.74 & 481s\\
Full Potential Target Mining     & \textbf{89.41} & \textbf{92.13} & \textbf{98.25} & \textbf{95.09} & \textbf{71.28} & \textbf{82.74} & \textbf{87.69} & \textbf{85.14} & 651s\\ \bottomrule
\end{tabular}}
\end{table*}

\subsubsection{Impacts of Pseudo-label Contrastive Learning}
\begin{figure}[h]
    \centering
\includegraphics[width=0.8\linewidth]{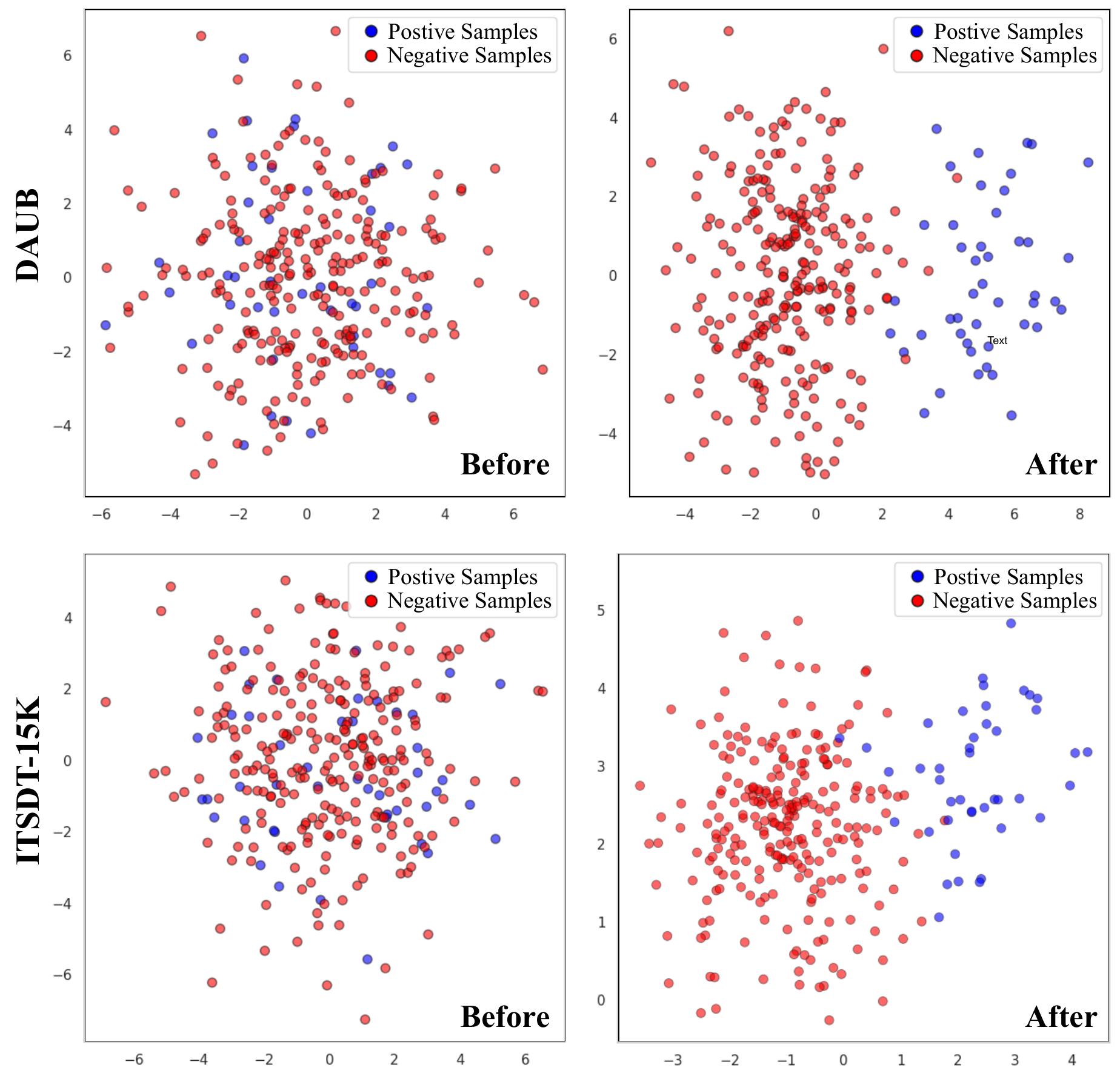}
    \caption{The feature distribution comparisons of pseudo-labels before and after our proposed PCL.}
    \label{fig:pcl}
\end{figure}

To visualize the effect of pseudo-label contrastive learning, we perform a group of feature distribution comparisons of pseudo-labels before and after PCL, as shown in Figure \ref{fig:pcl}.
From it, we could find that PCL could effectively separate positive and negative pseudo-labels on both DAUB and ITSDT-15K.
For example, on DAUB, the feature distributions of positive and negative samples are mixed together before PCL. However, after applying PCL, a clear distinction emerges between all the samples. The distance between positive samples decreases, while the distance between positive and negative samples increases.
It indicates that our PCL could enhance the model's ability to distinguish high-quality pseudo-labels. 

\subsubsection{Impacts of Motion-aware Learning}
\begin{figure}[h]
    \centering
    \includegraphics[width=\linewidth]{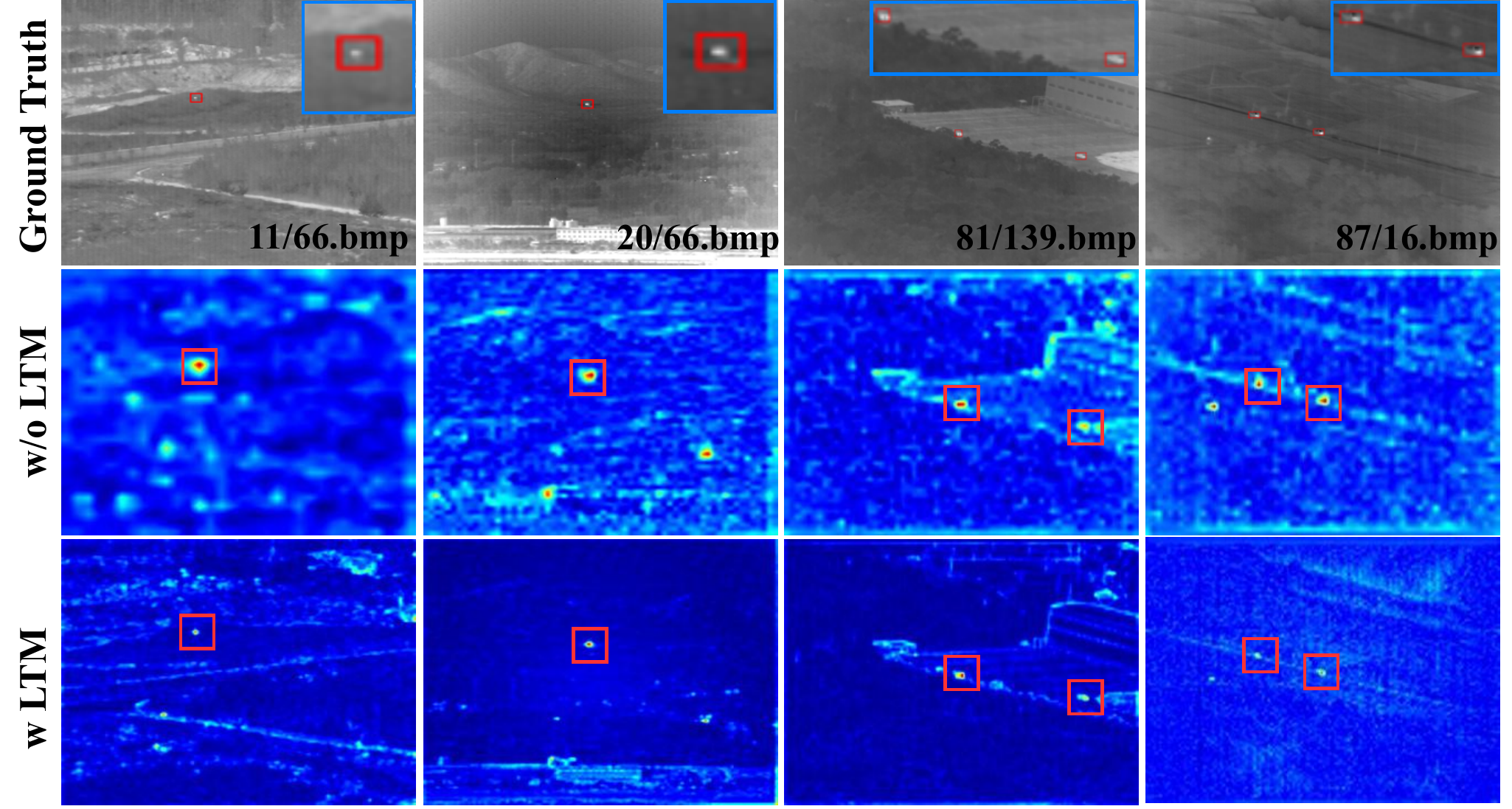}
    \caption{The feature heatmaps without (w/o) and with (w) LTM. The first two columns are on DAUB, and the last two are on ITSDT-15K.}
    \label{fig:heat}
\end{figure}

To further explore the effectiveness of motion-aware learning, we select two samples from DAUB and ITSDT-15K to visualize the feature heatmaps with and without LTM, as illustrated in Figure \ref{fig:heat}.
In this figure, it is apparent that the focus positions of feature heatmaps by ``w/o LTM" are obscure and targets are even lost in complex backgrounds. In contrast, after employing LTM (``w LTM''), the feature response of moving infrared small targets is effectively enhanced, in the meanwhile, the noisy background becomes clearer.
This indicates that our long-short term motion-aware learning could capture the comprehensive motion features of targets, integrating local motion patterns and global motion trajectory. 

\subsubsection{Effects of $\eta$ and $\gamma$ in Loss}
Regarding the two hyper-parameters $\eta$ and $\gamma$ in the total loss of our WeCoL, $i.e.$, $ \mathcal{L} = \eta\mathcal{L}_{det} + \gamma\mathcal{L}_{pcl}$, we conduct a group of ablation studies, as shown in Table \ref{tab:hyper}. 
From this table, we could observe that varying the values of $\eta$ and $\gamma$ significantly influences the model's performance. 
Besides, an optimal balance between the detection loss $\mathcal{L}_{det}$ and pseudo-label contrastive learning loss $\mathcal{L}_{pcl}$ is achieved through $\eta = 1$ and $\gamma = 1$.
\begin{table}[h]
\centering
\caption{Ablation study on the hyper-parameters $\eta$ and $\gamma$ in loss.}
\resizebox{\linewidth}{!}{
\begin{tabular}{c|cccccccc}
\toprule
\multirow{2}{*}{\textbf{($\eta$, $\gamma$)}} & \multicolumn{4}{c}{\textbf{DAUB}}                              & \multicolumn{4}{c}{\textbf{ITSDT-15K}}                         \\ \cmidrule{2-9}
                                & $\textbf{mAP}_{\textbf{50}}$ & \textbf{Pr}    & \textbf{Re} & \textbf{F1}    & $\textbf{mAP}_{\textbf{50}}$ & \textbf{Pr} & \textbf{Re}    & \textbf{F1}    \\ \midrule 
(1, 3) & 85.94 & 88.46 & \textbf{98.69} & 93.30 & 70.06 & 81.23          & 84.79 & 82.97 \\
(1, 2) & 82.43 & 85.81 & 97.14          & 91.13 & 68.74 & \textbf{83.51} & 85.39 & 84.44 \\
(1, 1)                          & \textbf{89.41} & \textbf{92.13} & 98.25       & \textbf{95.09} & \textbf{71.28} & 82.74       & \textbf{87.69} & \textbf{85.14} \\
(2, 1) & 88.28 & 90.96 & 98.67          & 94.66 & 70.47 & 81.59          & 86.35 & 83.90 \\
(3, 1) & 86.06 & 89.58 & 97.60          & 93.42 & 69.23 & 80.27          & 86.16 & 83.11 \\ \bottomrule
\end{tabular}}
\label{tab:hyper}
\end{table}

\subsubsection{Effects of $n$ in Potential Target Mining}
\begin{figure}[h]
    \centering
\includegraphics[width=\linewidth]{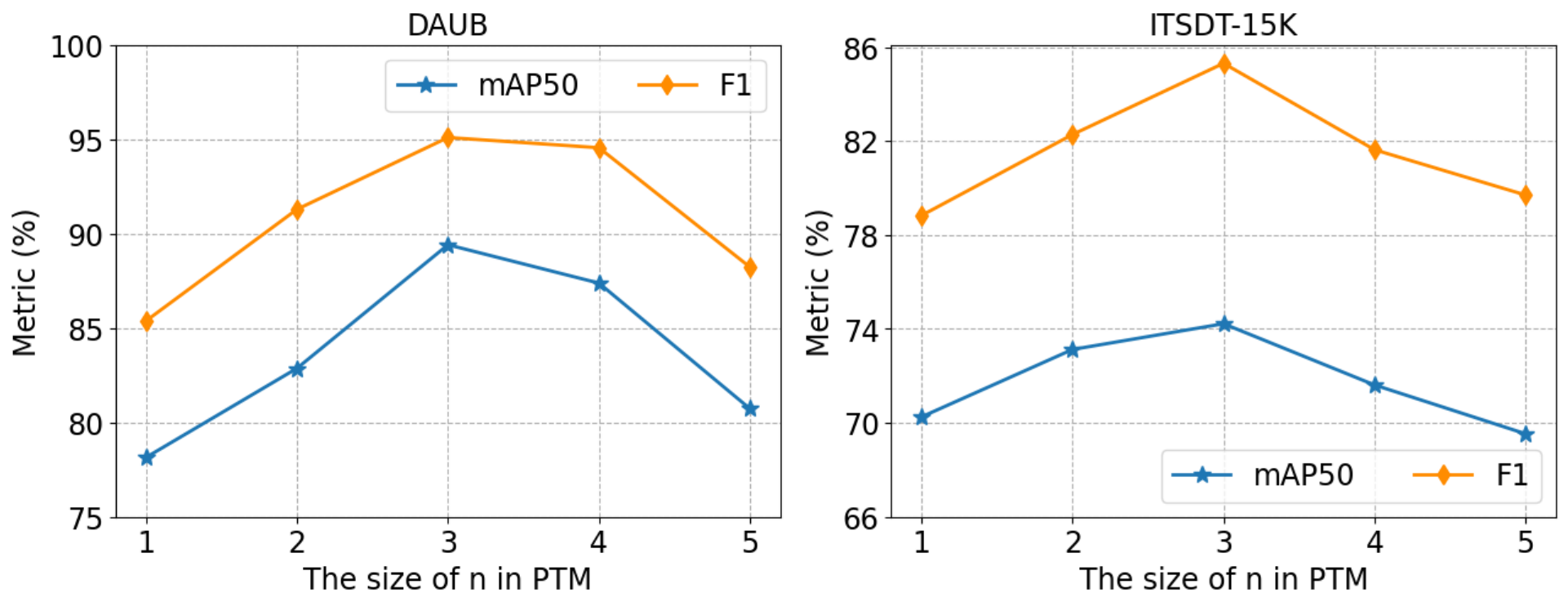}
    \caption{The effects of hyper-parameter $n$ in PTM on our WeCoL.}
    \label{fig:n}
\end{figure}

To investigate the effects of hyper-parameter $n$ in Eq. (3) on our proposed PTM, we conduct a group of ablation studies, as shown in Figure \ref{fig:n}.
Actually, $n$ is used to control the number of point prompts for generating pseudo-labels.
From the figure, it could be easily observed that the performance peak is achieved when $n$ is set to 3. 
This further indicates that our WeCoL could perform best when the number of point prompts is 3 times larger than the target quantity.
One possible reason is that if $n$ is too small, we could not obtain enough pseudo-labels for detector training. On the contrary, if $n$ is too big, the generated numerous pseudo-labels will contain too many errors.
In view of this, the optimal $n$ of our potential target mining seems to be 3 on two datasets. 

\subsection{Effects of Frame Number $T$}
\begin{figure}[h]
    \centering
\includegraphics[width=\linewidth]{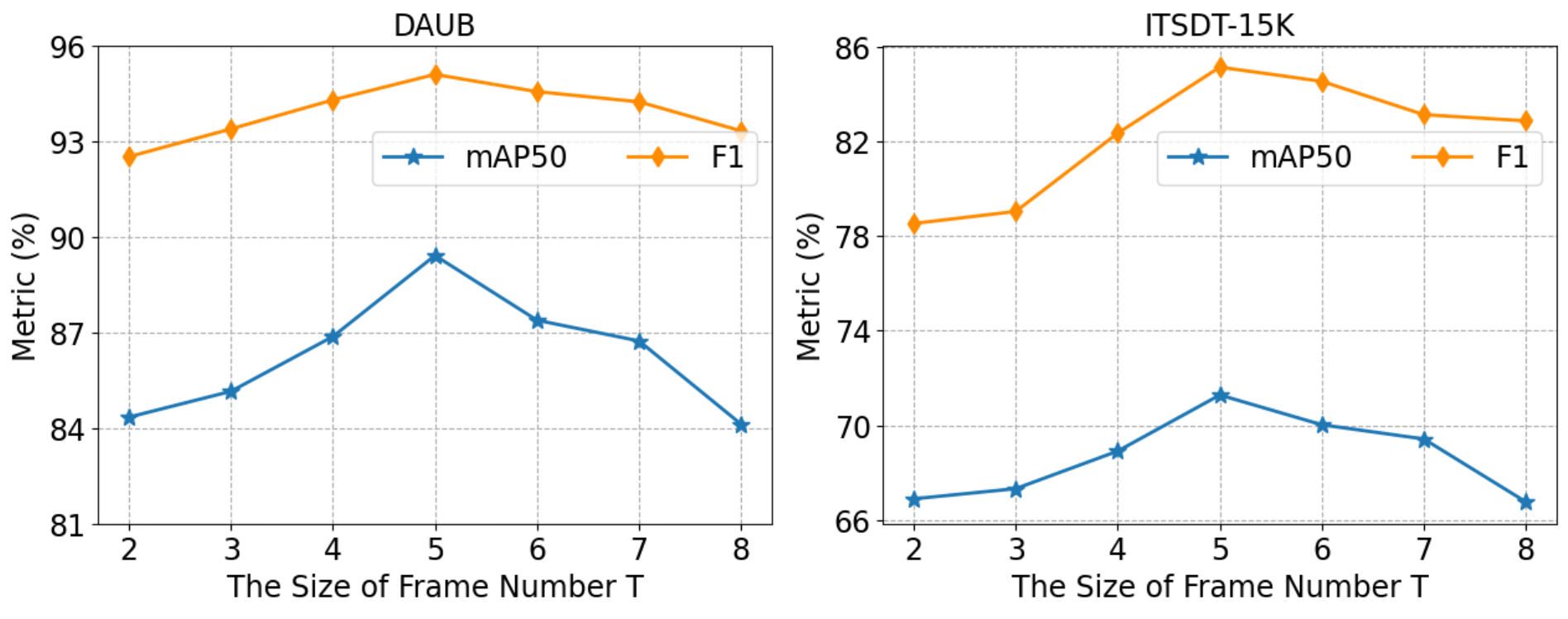}
    \caption{The effects of frame sampling size $T$ on our WeCoL.}
    \label{fig:number}
\end{figure}

To explore the effects of frame sampling number $T$ on detection performance, we perform a group of ablation studies, as shown in Figure \ref{fig:number}.
It is apparent that our WeCoL reaches the highest $\text{mAP}_{50}$ and F1 on both DAUB and ITSDT-15K when $T = 5$. 
One possible reason is that if $T$ is too small, the motion features of targets could not be captured effectively. On the contrary, if $T$ is too big, excessive redundant features could cause interference. 
Therefore, the optimal frame sampling number of our WeCoL seems to be 5 on two datasets. 

\section{Conclusion}
In this paper, we propose the first weakly-supervised contrastive learning framework, $i.e.$, WeCoL, to avoid the expensive target-wise manual annotations for MISTD. 
In detail, it utilizes SAM to generate reliable pseudo-labels through activation maps and multi-frame energy accumulation, so as to reduce pseudo-label redundancy.
Besides, weakly-supervised contrastive learning could further employ target quantity prompts to choose high-quality pseudo-labels, facilitating detector training.   
Extensive experiments verify the effectiveness and superiority of our WeCoL in moving infrared small target detection. On primary performance metrics, it could even reach to the performance over the 90\% of SOTA fully-supervised  methods. 
Its main weakness lies in its high model complexity and heavy reliance on pseudo-label quality.
In the future, an optimized lightweight weakly-supervised detection scheme with more accurate pseudo-label generation is worthy of further exploration.

\bibliographystyle{IEEEtran}
\bibliography{acmart} 

\end{document}